%% file: acl_latex.tex
\pdfoutput=1

\documentclass[11pt]{article}

\usepackage[]{acl}

\usepackage{times}
\usepackage{latexsym}

\usepackage[T1]{fontenc}

\usepackage[utf8]{inputenc}
\usepackage{algorithm}
\usepackage{algorithmic}

\usepackage{microtype}
\usepackage{boldline}
\usepackage{multirow}
\usepackage{makecell}
\usepackage{hhline}
\usepackage{pgfplots}
\usepackage{tikz}
\usepackage{enumitem}

\usepackage{mathtools}
\usepackage{pifont}

\newcommand{\mcL}{\mathcal{L}}

\newcommand{\mcY}{\mathcal{Y}}

\newcommand{\xmark}{\textcolor{red}{\ding{54}}}
\input{math_commands.tex}

\title{ITA: Image-Text Alignments for Multi-Modal Named Entity Recognition}

\author{Xinyu Wang$^{\diamondsuit\ddagger}$, Min Gui$^{\heartsuit}$, Yong Jiang$^{\spadesuit}$\textsuperscript{$\ast$}, Zixia Jia$^{\diamondsuit}$, Nguyen Bach$^{\clubsuit}$, Tao Wang,\\
\textbf{Zhongqiang Huang$^{\spadesuit}$, Fei Huang$^{\spadesuit}$,  Kewei Tu$^{\diamondsuit}$}\thanks{\hspace{1mm} Yong Jiang and Kewei Tu are the corresponding authors. $^{\ddagger}$: This work was done when Xinyu Wang, Min Gui and Nguyen Bach were at Alibaba Group. } \\
 $^\diamondsuit$School of Information Science and Technology, ShanghaiTech University \\
 $^\diamondsuit$Shanghai Engineering Research Center of Intelligent Vision and Imaging \\
 $^\spadesuit$DAMO Academy, Alibaba Group \\
 $^{\heartsuit}$Shopee, Singapore \\
 $^{\clubsuit}$Microsoft \\
  {\tt \{wangxy1,jiazx,tukw\}@shanghaitech.edu.cn, min.gui@shopee.com} \\
  {\tt \{yongjiang.jy,z.huang,f.huang\}@alibaba-inc.com} \\
  {\tt nguyenbach@microsoft.com} \\
}

\begin{document}
\maketitle

\begin{abstract}
Recently, Multi-modal Named Entity Recognition (MNER) has attracted a lot of attention. Most of the work utilizes image information through region-level visual representations obtained from a pretrained object detector and  relies on an attention mechanism to model the interactions between image and text representations. However, it is difficult to model such interactions as image and text representations are trained separately on the data of their respective modality and are not aligned in the same space. As text representations take the most important role in MNER, in this paper, we propose {\bf I}mage-{\bf t}ext {\bf A}lignments (ITA) to  align image features into the textual space, so that the attention mechanism in transformer-based pretrained textual embeddings can be better utilized. ITA first aligns the image into regional object tags, image-level captions and optical characters as visual contexts, concatenates them with the input texts as a new cross-modal input, and then feeds it into a pretrained textual embedding model. This makes it easier for the attention module of a pretrained textual embedding model to model the interaction between the two modalities since they are both represented in the textual space. ITA further aligns the output distributions predicted from the cross-modal input and textual input views so that the MNER model can be more practical in dealing with text-only inputs and robust to noises from images. In our experiments, we show that ITA models can achieve state-of-the-art accuracy on multi-modal Named Entity Recognition datasets, even without image information.\footnote{Our code is publicly available at \url{https://github.com/Alibaba-NLP/KB-NER/tree/main/ITA}.}


\end{abstract}

\section{Introduction}
Named Entity Recognition (NER) \cite{Sundheim1995NamedET} has attracted increasing attention in natural language processing community. It has been applied to a lot of domains such as news \citep{tjong-kim-sang-2002-introduction,tjong-kim-sang-de-meulder-2003-introduction}, E-commerce \citep{10.1145/3404835.3463102}, social media \cite{strauss-etal-2016-results,derczynski-etal-2017-results} and bio-medicine \citep{dougan2014ncbi,li2016biocreative}. Several recent studies focus on improving the accuracy of NER models through utilizing image information (MNER) in tweets \citep{zhang2018adaptive,moon-etal-2018-multimodal,lu-etal-2018-visual}. Most approaches to MNER use the attention mechanism to model the interaction between image and text representations \citep{yu-etal-2020-improving-multimodal,zhang2021multi,Sun2021RpBERTAT}, in which image representations are from a pretrained feature extractor, i.e. ResNet \citep{he2016deep}, and text representations are extracted from pretrained textual embeddings, i.e. BERT \citep{devlin-etal-2019-bert}. Since these models are separately trained on datasets of different modalities and their feature representations are not aligned, it is difficult for the attention mechanism to model the interaction between the two modalities. 


Recently, pretrained vision-language (V+L) models such as LXMERT \citep{Tan2019LXMERTLC}, UNITER \citep{chen2020uniter} and Oscar \citep{li2020oscar} have achieved significant improvement on several cross-modal tasks such as image captioning, VQA \citep{Agrawal2015VQAVQ}, NLVR \citep{Young2014FromID} and image-text retrieval \citep{Suhr2019ACF}. Most pretrained V+L models are trained on image-text pairs and simply concatenate text features and image features as the input of pretraining. There are, however, two problems. First, texts in these datasets mainly contain common nouns instead of named entities\footnote{\url{https://visualgenome.org/data_analysis/statistics}} which leads to an inductive bias over common nouns and images. Second, despite its important role in pretraining V+L models,  the image modality only plays an auxiliary role in MNER for disambiguation, and can sometimes even be discarded. These problems make pretrained V+L models perform weaker than pretrained language models for MNER.

Pretrained textual embeddings such as BERT, XLM-RoBERTa \citep{conneau-etal-2020-unsupervised} and LUKE \citep{yamada-etal-2020-luke} have achieved state-of-the-art performance on various NER datasets through simple fine-tuning of pretrained textual embeddings. Since most of the transformer-based pretrained textual embeddings are trained over long texts, recent work \citep{akbik-etal-2019-pooled,schweter2020flert,yamada-etal-2020-luke,wang-etal-2021-improving} has shown that introducing document-level contexts can significantly improve the accuracy of a NER model. The attention mechanism in transformer-based pretrained textual embeddings can utilize contexts to improve the token representation of a sequence. Moreover, pretrained V+L models such as Oscar and VinVL \citep{zhang2021vinvl} can use object tags detected in images to significantly ease the alignments between text and image features. Therefore, the images in MNER can be converted to texts as well so that the image representations can be aligned to the space of text representations. As a result, the attention module of the pretrained textual embeddings have the capability to easily model the interactions between aligned image and text representations, without introducing a new attention module. In this paper, we propose ITA, a simple but effective framework for {\bf I}mage-{\bf T}ext {\bf A}lignments. ITA converts an image into visual contexts in textual space by multi-level alignments. We concatenate the NER texts with the visual contexts as a new cross-modal input view and then feed it into a pretrained textual embedding model to improve the token representations of NER texts, which are fed into a linear-chain CRF \citep{10.5555/645530.655813} layer for prediction.
In practice, a MNER model should be robust when there is only text information, as images may be unavailable or can introduce noises. Sometimes it is even undesirable to use images as image feature extraction can be inefficient in online serving. Therefore, we further propose to utilize the cross-modal input view to improve the accuracy of textual input view, based on cross-view alignment that minimizes the KL divergence over the probability distributions of the two views. 

ITA can be summarized in four aspects:
\begin{enumerate}[leftmargin=*]
    \item Object Tags as Local Alignment: ITA locally extracts object tags and its corresponding attributes of image regions from an object detector.
    \item Image Captions as Global Alignment: ITA summarizes what the image is describing through predicting image captions from an image captionilng model.
    \item Optical Character Alignment: ITA extracts the texts presented in the image via optical character recognition (OCR).
    \item Cross-View Alignment: we calculate the KL divergence between the output distributions of two input views.
\end{enumerate}
We show in experiments that ITA can significantly improve the model accuracy on MNER datasets and achieve the state-of-the-art. The cross-view alignment module can significantly improve both the cross-modal and textual input views, and bridge the performance gap between the two views. 

\begin{figure*}[ht]
	\centering
	\includegraphics[scale=0.61]{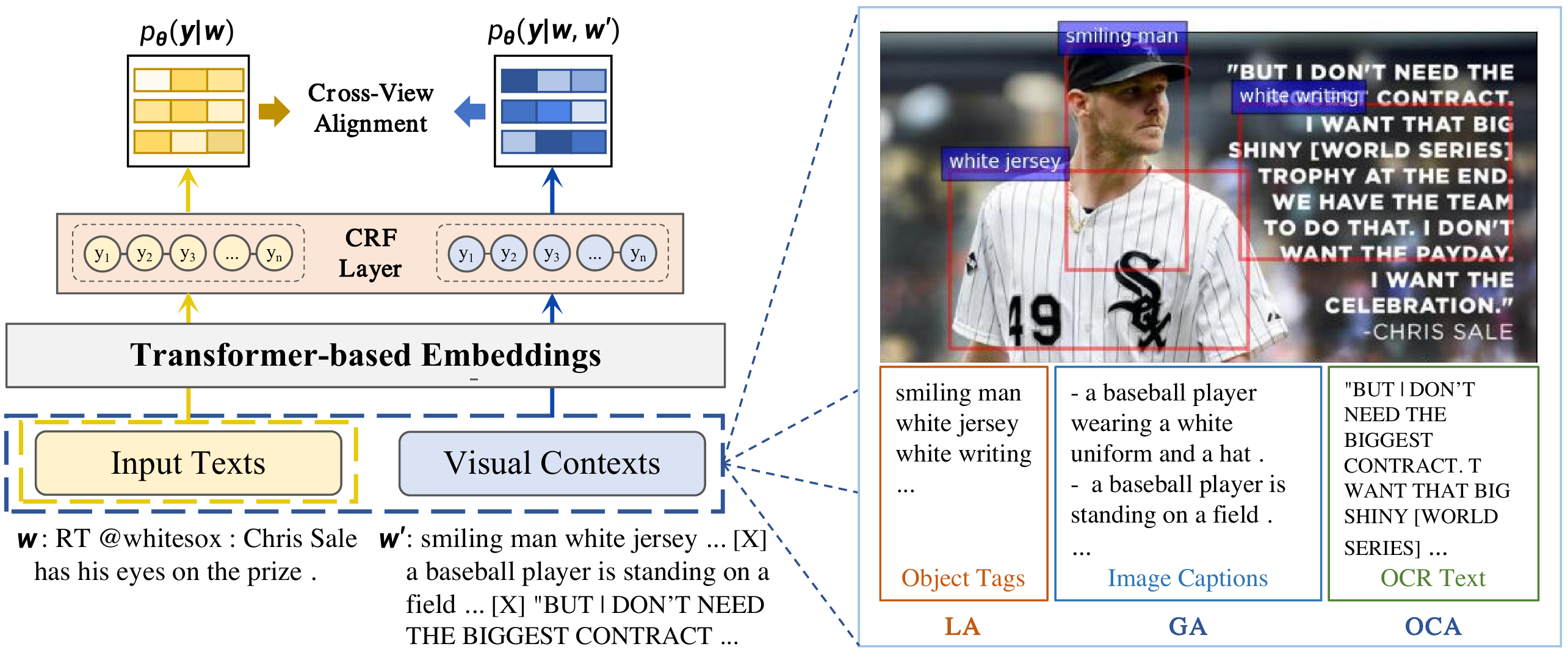}
	\caption{The architecture of ITA. ITA aligns an image into object tags, image captions and texts from OCR. ITA takes them as visual contexts and then feeds them together with the input texts into the transformer-based embeddings. In the cross-view alignment module, ITA minimizes the distance between the output distribution of cross-modal inputs and textual inputs. 
	}
	\label{fig:architecture}
\end{figure*}

\section{Approaches}
We consider the NER task as a sequence labeling problem. Given a sentence $\vw = \{w_1, \cdots, w_n\}$ with $n$ tokens and its corresponding image $I$, an sequence labeling model aims to predict a label sequence $\vy = \{y_1, \cdots, y_n\}$ at each position.
In our framework, we focus on incorporating visual information to improve the representations of the input tokens by aligning visual and textual information effectively. We use a visual context generator to convert the image $I$ into texts forming visual contexts $\vw^\prime = \{w_1^\prime, \cdots, w_m^\prime\}$ with $m$ tokens. We then concatenate the input text and visual contexts as a cross-modal text+image (\textbf{I+T}) input view instead of the text (\textbf{T}) input view. We feed the \textbf{I+T} input into a pretrained textual embeddings model to get stronger token representations of the input sentence. Then the token representations are fed into a linear-chain CRF layer to get the label sequence $\vy$. To further improve the model accuracy of both input views, we use the cross-view alignment module to align the output distributions of \textbf{I+T} and \textbf{T} input views during training. The architecture of our framework is shown in Figure \ref{fig:architecture}.

\subsection{NER Model Architecture}
We use a neural model with a linear-chain CRF layer, a widely used approach for the sequence labeling problem \citep{Huang2015BidirectionalLM,akbik-etal-2018-contextual,devlin-etal-2019-bert}. The input is fed into a transformer-based pretrained textual embeddings model and the output token representations $\{\vr_1, \cdots, \vr_n\}$ are fed into the CRF layer:
\begin{align}
    p_\theta(\vy|\vw) &= \frac{\prod\limits_{i=1}^{n} \psi(y_{i-1}, y_i, \vr_i)}{\sum\limits_{\vy' \in \mathcal{Y}(\vw)} \prod\limits_{i=1}^{n} \psi(y'_{i-1}, y'_i, \vr_i)}\nonumber
\end{align}
where $\theta$ is the model parameters, $\mathcal{Y}(\vw)$ is the set of all possible label sequences given the input $\vw$. Given the gold label sequence $\hat{\vy}$ in the training data, the objective function of the model for the \textbf{T} input view is:
\begin{align}
\mcL_\text{T}(\theta) = - \log p_\theta(\hat{\vy}|\vw) \label{eq:nll_loss}
\end{align}
The loss can be calculated using Forward algorithm.

\subsection{Image-text Alignments}
The transformer-based pretrained textual embeddings have strong representations over texts. Therefore, ITA converts the image information into textual space through generating texts from the image so that the learning of the self-attention in the transformer-based model can be significantly eased compared with simply using image features from an object detector. We propose a local (LA), a global (GA) and an optical character alignment (OCA) approaches for alignments.
\paragraph{Object Tags as Local Alignment} Given an image, the image information can be decomposed into a set of objects in local regions. The object tags of each region textually describe the local information in the image. To extract the objects, we use an object detector $\textbf{OD}$ to identify and locate the objects in the image: 
\begin{align}
&\va, \vo = \textbf{OD}(I); \text{where }  \nonumber\\
&\va = \{\va_1,\va_2,\cdots,\va_l\} \text{ and } \vo = \{\evo_1,\evo_2,\cdots,\evo_l\} \nonumber
\end{align}
The attribute predictions from the object detector contain multiple attribute tags $\va_i$ for each object $\evo_i$. We linearize and sort the objects in a descending order based on the confidences of the detection model. For each object, we heuristically keep 0 to 3 attributes with confidence scores above a threshold $m$. We linearize the attributes and put the attributes before the corresponding objects since the attributes are the adjectives describing the object tags. 
As a result, we take the predicted $l$ object tags $\vo$ and their attribute tags $\va$ from the object detector as the locally aligned visual contexts $\vw^{\text{LA}}$:
\begin{align}
\vw^{\text{LA}} = \{\va_1,\evo_1,\va_2,\evo_2,\cdots,\va_l,\evo_l\}\nonumber
\end{align}

\paragraph{Image Captions as Global Alignment} Though the local alignment can localize the image into objects, the objects cannot fully describe the of the whole image. Image captioning is a task that predicts the meaning of an image. Therefore, we align the image into $k$ image captions by an image captioning model $\textbf{IC}$:
\begin{align}
\{\vw^1,\vw^2,\cdots,\vw^k\} = \textbf{IC}(I) \nonumber
\end{align}
where $\{\vw^1,\vw^2,\cdots,\vw^k\}$ are captions generated from beam search with $k$ beams. We concatenate the $k$ captions together with a special separate token [X] to form the aligned global visual contexts $\vw^{\text{GA}}$:
\begin{align}
 \vw^{\text{GA}}  {=} [\vw^1,\text{[X]},\vw^2,\text{[X]},\cdots,\text{[X]},\vw^k] \nonumber
\end{align}
The exact label (e.g. ``[SEP]'' in BERT) of the special [X] token depends on the selection of embeddings.

\paragraph{Optical Character Alignment} Some image contain text when they are created to enrich the semantic information that the images want to convey. In order to better understand this type of image, we use an \textbf{OCR} model to identify and extract the texts in the image:
\begin{align}
\vw^{\text{OCA}} = \textbf{OCR}(I) \nonumber
\end{align}
where $\vw^{\text{OCA}}$ are the texts extracted by the \textbf{OCR} model. Note that $\vw^{\text{OCA}}$ may be an empty text if there is no text in the image.

We concatenate the input sentence and our aligned visual contexts to form the \textbf{I+T} input view $\hat{\vw}=[\vw;\vw^\prime]$, where $\vw^\prime$ can be one of $\vw^{\text{LA}}$, $\vw^{\text{GA}}$, $\vw^{\text{OCA}}$ or the concatenation of all (we denote it as \textbf{All}). The transformer-based embeddings are fed with the \textbf{I+T} input view and then output image-text fused token representations for each token $\{\vr_1^\prime, \cdots, \vr_n^\prime\}$. The token representations are fed into the CRF layer to get the probability distribution $p_\theta(\vy|\hat{\vw})$. Similar to Eq. \ref{eq:nll_loss}, the objective function of the model for the \textbf{I+T} input view is:
\begin{align}
\mcL_{\text{I+T}}(\theta) = - \log p_\theta(\hat{\vy}|\hat{\vw}) \label{eq:nll_x}
\end{align}
\paragraph{Cross-View Alignment}
There are several limitations in incorporating images into NER prediction: 1) the images may not available in testing; 2) aligning images to texts requires several pipelines in pre-processing instead of an end-to-end manner, which is so time-consuming that it is not applicable to some time-critical scenes such as online serving; 3) the noises in the image can mislead the MNER model to make wrong predictions. To alleviate these issues, we propose Cross-View Alignment (CVA), which targets at reducing the gap between the \textbf{I+T} and \textbf{T} input views over the output distributions so that the MNER model can better utilize the textual information in the input.
During training, CVA minimizes the KL divergence over the probability distribution of \textbf{I+T} and \textbf{T} input views:
\begin{align}
\mcL_{\text{CVA}}(\theta)&{=}\text{KL}(p_\theta(\vy|\hat{\vw})||p_\theta(\vy|\vw)) \label{eq:mda_loss}
\end{align}
Since the \textbf{I+T} input view has additional visual information in the input and we want the \textbf{T} input view to match the accuracy of \textbf{I+T} input view, we only back-propagate through $p_\theta(\vy|\vw)$ in Eq. \ref{eq:mda_loss}. Therefore, Eq. \ref{eq:mda_loss} is equivalent to calculating the cross-entropy loss over the two distributions:
\begin{align}
\mcL_{\text{CVA}}(\theta)&{=}\sum_{\vy \in \mcY(\vx)} p_\theta(\vy|\hat{\vw}) \log p_\theta(\vy|\vw) \label{eq:xe_loss}
\end{align}
As the set of all possible label sequences $ \mcY(\vx)$ is exponential in size, we calculate the posterior distributions of each position $p_\theta(\evy_i|\vw)$ and $p_\theta(\evy_i|\hat{\vw})$ through forward-backward algorithm to approximate Eq. \ref{eq:xe_loss}:
\begin{align}
&p_\theta(\evy_k|*) {\propto} \sum\limits_{\{y_0,\dots,y_{k-1}\}} {\prod\limits_{i=1}^{k}} \psi(y_{i-1}, y_i, \vr^*_i) \nonumber\\
&{\times} \sum\limits_{\{y_{k{+}1},\dots,y_n\}} {\prod\limits_{i{=}k{+}1}^{n}} \psi(y_{i{-}1}, y_i, \vr^*_i) \nonumber \\
&\mcL_{\text{CVA}}(\theta){=}\sum_{i=1}^n p_\theta(\evy_i|\hat{\vw}) \log p_\theta(\evy_i|\vw)) \label{eq:mda} 
\end{align}
where $\vr^*_i$ represents either $\vr_i$ or $\vr_i^\prime$.
\paragraph{Training} During training, we jointly train \textbf{T} and \textbf{I+T} input views with the training objective in Eq. \ref{eq:nll_loss} and \ref{eq:nll_x} together with the CVA alignment training objective in Eq. \ref{eq:mda}. As a result, the final training objective for ITA is:
\begin{align}
\mcL_{\text{ITA}} = \mcL_{\text{CVA}} + \mcL_{\text{T}} + \mcL_{\text{I+T}} \nonumber
\end{align}




\section{Experiments}
\label{sec:exp}
We conduct experiments on three MNER datasets. To show the effectiveness of our approaches, we use two embedding settings and compare our approaches with previous multi-modal approaches.
\subsection{Settings}

\paragraph{Datasets}
We show the effectiveness of our approaches on Twitter-15, Twitter-17 and SNAP Twitter datasets\footnote{Twitter-15 and 17 datasets are available at \url{https://github.com/jefferyYu/UMT}.} containing 4,000/1,000/3,357, 3,373/723/723 and 4,290/1,432/1,459 sentences in train/development/test split respectively. The Twitter-15 dataset is constructed by \citet{zhang2018adaptive}. The SNAP dataset is constructed by \citet{lu-etal-2018-visual} and the Twitter-17 dataset is a filtered version of SNAP constructed by \citet{yu-etal-2020-improving-multimodal}.

\paragraph{Model Configuration}
For token representations, we use BERT base model to fairly compare with most of the recent work \citep{yu-etal-2020-improving-multimodal,zhang2021multi,Sun2021RpBERTAT}. Recently, XLM-RoBERTa has achieved state-of-the-art accuracy on various NER datasets by feeding the input together with contexts to the model. To further utilize the visual contexts in transformer-based embeddings, we use XLM-RoBERTa large (XLMR) model as another embedding in our experiments. 
To extract object tags and image captions of the image, we use VinVL \citep{zhang2021vinvl}, which is a pretrained V+L model based on a newly pretrained large-scale object detector based on the ResNeXt-152 C4 architecture. We use the object detection module of VinVL to predict object tags and their corresponding attributes. The number of object tags and attributes varies over the images and is no more than $100$. We set the threshold $m$ to be $0.1$ for keeping the attributes of each object. For image captions, we use VinVL large model finetuned on MS-COCO \citep{lin2014microsoft} captions\footnote{\url{github.com/microsoft/Oscar}} with CIDEr optimization \citep{rennie2017self}. In our experiments, we use a beam size of $5$ with at most 20 tokens for prediction and keep all the $5$ captions as the visual contexts. For OCR, we use Tesseract OCR\footnote{\url{github.com/tesseract-ocr/tesseract}} \citep{smith2007overview}, which is an open source OCR engine. We use the default configuration of the engine to extract texts in the image\footnote{Please refer to Appendix \ref{app:oca} for more statistics.}.

\paragraph{Training Configuration}
During training, we finetune the pretrained textual embedding model by AdamW \citep{loshchilov2018decoupled} optimizer. In experiments we use the grid search to find the learning rate for the embeddings within $[1\times 10^{-6}, 5\times 10^{-4}]$. For BERT embeddings, we finetune the embeddings with a learning rate of $5\times 10^{-5}$ with a batch size of $16$. For XLMR embeddings, we use a learning rate of $5\times 10^{-6}$ and a batch size of $4$ instead. For the learning rate of the CRF layer, we use a grid search over $[0.05,0.5]$ and $[0.005,0.05]$ for BERT and XLMR respectively. The MNER models are trained for $10$ epochs and we report the average results from $5$ runs with different random seeds for each setting.

\begin{table}[t!]
\small
\centering
\setlength\tabcolsep{1pt}
\begin{tabular}{c|l||cc|cc|cc}
\hlineB{4}
\multicolumn{2}{c||}{} & \multicolumn{2}{c|}{Twitter-15} & \multicolumn{2}{c|}{Twitter-17} & \multicolumn{2}{c}{SNAP} \\
 \hline
\multirow{3}{*}{\makecell{Train\\Modal}} & \multirow{3}{*}{Approach} & \multicolumn{2}{c|}{\bf Eval} & \multicolumn{2}{c|}{\bf Eval} & \multicolumn{2}{c}{\bf Eval}\\
 & & \multicolumn{2}{c|}{\bf Modal} & \multicolumn{2}{c|}{\bf Modal} & \multicolumn{2}{c}{\bf Modal}\\
&  & {\textbf{T}} & {\textbf{I+T}} & {\textbf{T}} & {\textbf{I+T}} & {\textbf{T}} & {\textbf{I+T}}\\
\hline\hline
 \multicolumn{8}{c}{\bf \textsc{BERT-CRF}}\\
\hline
T & {\textbf{BERT-CRF}} & 74.79 & - & 85.18 & - & 85.98 & -\\
\hline
\multirow{7}{*}{I+T} & {\textbf{ITA-LA}} & - & 75.18 & - & 85.67& - & 86.26\\
 & {\textbf{ITA-GA}} & - & 75.17  & - & 85.75  & - & 86.72\\
 & {\textbf{ITA-OCA}} & - & 75.01 & - & 85.64& - & 86.52\\
 & {\textbf{ITA-All}} & - & 75.15 & - & 85.78 & - & 86.79\\
 & {\textbf{ITA-LA$_{\text{+CVA}}$}} & 75.26 & 75.20 & 85.72 & 85.62& 86.51 & 86.41\\
 & {\textbf{ITA-GA$_{\text{+CVA}}$}} & 75.45 & 75.52 & 85.96 & \textbf{85.85}& 86.42 & 86.39\\
 & {\textbf{ITA-OCA$_{\text{+CVA}}$}} & 75.26 & 75.30 & 85.73 & 85.79& 86.64 & 86.59\\
 & {\textbf{ITA-All$_{\text{+CVA}}$}} & \textbf{75.67} & \textbf{75.60} & \textbf{85.98} & 85.72&  \textbf{86.83} & \textbf{86.75} \\
\hline
\multicolumn{8}{c}{\bf \textsc{XLMR-CRF}}\\
\hline
T & {\textbf{XLMR-CRF}} & 77.37 & - & 88.73 & - & 89.39 & - \\
\hline
\multirow{7}{*}{I+T} & {\textbf{ITA-LA}} & - & 77.64 & - & 89.29& - & 89.68\\
 & {\textbf{ITA-GA}} & - & 77.78 & - & 89.32& - & 89.78 \\
& {\textbf{ITA-OCA}} & - & 77.94 & - & 89.31& - & 89.64\\
& {\textbf{ITA-All}} & - & 77.81 & - & 89.62 & - & 90.10\\
& {\textbf{ITA-LA$_{\text{+CVA}}$}} & 77.87 & 77.93 & 89.45 & \textbf{89.90}& 89.85 & 89.91\\
 & {\textbf{ITA-GA$_{\text{+CVA}}$}} & 78.03 & 78.02 & 89.41 & 89.62& 89.85 & 90.09 \\
& {\textbf{ITA-OCA$_{\text{+CVA}}$}} & 77.57 & 77.59 & 89.32 &89.55& 89.90 & 89.84\\
& {\textbf{ITA-All$_{\text{+CVA}}$}} & \textbf{78.25} & \textbf{78.03} & \textbf{89.47} & 89.75& \textbf{90.02} & \textbf{90.15}\\
\hlineB{4}
\end{tabular}
\caption{A comparison of ITA and our baseline.}
\label{tab:main}
\end{table}

\begin{table}[t!]
\small
\centering
\begin{tabular}{l||c|c|c}
\hlineB{4}
\multicolumn{1}{c||}{Approach} & \multicolumn{1}{c|}{Twitter-15} & \multicolumn{1}{c|}{Twitter-17} & \multicolumn{1}{c}{SNAP} \\
\hline
\multicolumn{4}{c}{\bf \textsc{Reported F1 of Previous Approaches}}\\
\hline
{\textbf{BERT-CRF}}$^\dagger$ & 71.81 &  83.44 & -\\
{\textbf{OCSGA}}$^\clubsuit$ & 72.92 & - & - \\
{\textbf{UMT}}$^\dagger$ & 73.41 & 85.31 & -\\
{\textbf{RIVA}}$^\ddagger$ & 73.80 & - & 86.80 \\ 
{\textbf{RpBERT$_{\text{base}}$}}$^\spadesuit$ &74.40 & - & 87.40 \\
{\textbf{UMGF}}$^\diamond$ & 74.85 & 85.51 & - \\
\hline
\multicolumn{4}{c}{\bf \textsc{Our Reproductions}}\\
\hline
{\textbf{BERT-CRF}} & 74.79 & 85.18 & 85.98\\
{\textbf{UMT}} & 72.83 & 84.88 & - \\
{\textbf{UMGF}} & 74.42 & 85.27 & - \\ 
{\textbf{RpBERT$_{\text{base}}$}} & 67.21 & - & 62.14 \\ 
{Ours: \textbf{ITA-All$_{\text{+CVA}}$}} & \textbf{76.01} & \textbf{86.45} & \textbf{87.44} \\ 
\hlineB{4}
\end{tabular}
\caption{A comparison of our approaches and state-of-the-art approaches. $\clubsuit$: \citet{10.1145/3394171.3413650}; $\dagger$: results are from \citet{yu-etal-2020-improving-multimodal}; $\ddagger$: \citet{sun-etal-2020-riva}, $\spadesuit$: \citet{Sun2021RpBERTAT}, note that {\textbf{RpBERT$_{\text{base}}$}} uses the test set to select the best model; $\diamond$: results are from \citet{zhang2021multi}.}
\label{tab:previous_sota}
\end{table}

\subsection{Results}
In Table \ref{tab:main}, we compare our approaches with our baselines with different training and evaluation modalities (\textbf{T} for the text-only input view and \textbf{I+T} for the multi-modal input view). Results show that ITA models are significantly stronger than our {\textbf{BERT-CRF}} and {\textbf{XLMR-CRF}} baselines (Student's t-test with $p<0.05$). For the aligned visual contexts, LA, GA and OCA are competitive in most of the cases. To show the effectiveness of CVA, we report the evaluation results of both input views in evaluation. With CVA, the accuracy of both input views can be improved, especially the \textbf{T} input view. CVA can improve the \textbf{T} input view to be competitive with \textbf{I+T} input view. Moreover, the combination of all the alignments \textbf{ITA-All$_{\text{+CVA}}$} can further improve the model accuracy in most of the cases. The accuracy of the MNER models can be significantly improved if we use XLMR embeddings, which shows the importance of the text modality in MNER. With XLMR embeddings, the model accuracy can be further improved with ITA. The relative improvements over the baseline models are sometimes higher with XLMR than with BERT, which shows that the visual contexts can be further utilized with stronger embeddings. 

In Table \ref{tab:previous_sota}, we compare \textbf{ITA} with previous state-of-the-art approaches. For previous approaches, we report the results including
\textbf{OCSGA}, \textbf{UMT}, \textbf{RIVA}, \textbf{RpBERT}, \textbf{UMGF}, which are the proposed approaches of \citet{10.1145/3394171.3413650}, \citet{yu-etal-2020-improving-multimodal}, \citet{sun-etal-2020-riva}, \citet{Sun2021RpBERTAT} and \citet{zhang2021multi} respectively. For fair comparison, we report the results of these models based on the BERT base embeddings. Moreover, since most of these previous approaches report the best model accuracy instead of the averaged model accuracy, we use the best model accuracy of \textbf{ITA-All$_{\text{+CVA}}$} over $5$ runs. We also report our reproduced results of \textbf{UMT}, {\textbf{RpBERT}} and \textbf{UMGF} on the corresponding datasets. 
The results show that \textbf{ITA-All$_{\text{+CVA}}$} outperforms all of the previous approaches. On the SNAP dataset, the reported accuracy of {\textbf{RpBERT$_{\text{base}}$}} is competitive with \textbf{ITA-All$_{\text{+CVA}}$}. However, we find that the accuracy of our reproduced {\textbf{RpBERT$_{\text{base}}$}}\footnote{We reproduced the results based on the official code for {\textbf{RpBERT$_{\text{base}}$}}: \url{https://github.com/Multimodal-NER/RpBERT}} is significantly lower than the reported accuracy, even after careful check of the source code and hyper-parameter tuning.
Moreover, the fact that our \textbf{BERT-CRF} baseline achieves competitive accuracy with previous state-of-the-art multi-modal approaches shows that most of the previous work has not fully explored the strength of the text representations for the task.

\begin{table}[t!]
\small
\centering
\setlength\tabcolsep{4pt}
\begin{tabular}{l||c|c}
\hlineB{4}
Approaches & \multicolumn{1}{c|}{Twitter-15} & \multicolumn{1}{c|}{Twitter-17} \\
\hline\hline
{\textbf{BERT-CRF$_{\text{UMT}}$}} & 71.81 &  83.44 \\
{\textbf{BERT-CRF$_{\text{Ours}}$}} & 74.79 & 85.18 \\
\hline
\multicolumn{3}{c}{\bf \textsc{Our Reproductions}}\\
\hline
{\textbf{BERT-CRF$_{\text{UMT}}$}} & 71.74 & 84.20 \\
{\textbf{BERT-CRF$_{\text{UMT-Improved}}$}} & 72.53 & 84.48 \\
{\textbf{UMT}} & 72.83 & 84.88 \\
{\textbf{UMT$_{\text{Improved}}$}}  & 72.96 & 84.50 \\
\hlineB{4}
\end{tabular}
\caption{Our reproductions of previous baselines and approaches. ``Improved'' means our improved models based on the UMT code base.}
\label{tab:reproduction}
\end{table}

\paragraph{Discussion about Textual Modules}
As we have shown in Table \ref{tab:main} and \ref{tab:previous_sota}, the textual baselines (i.e. {\textbf{BERT-CRF}}) of previous work are significantly lower than that of ours. In most of the previous MNER architectures, the textual modules are mainly based on the baseline architectures with some modifications. We further show the baselines of previous work are not well-trained and how the multi-modal approaches perform with stronger textual modules. In Table \ref{tab:reproduction}, we rerun the \textbf{BERT-CRF} baseline based on the released codes of \textbf{UMT}\footnote{\url{https://github.com/jefferyYu/UMT}}. Based on the code of \textbf{UMT}, we tried to improve the baseline models in the code by using the same loss function as ours\footnote{The details are discussed in Appendix \ref{app:bug}}. The accuracy of \textbf{BERT-CRF} models in the code are significantly improved but the \textbf{UMT} models based on the improved code are not improved and even get worse in Twitter-17. Therefore, we suspect the \textbf{UMT} model cannot be further improved even with stronger textual modules. \citet{zhang2021multi} also reported the baseline based on the implementation of \citet{yu-etal-2020-improving-multimodal}, so we suspect the \textbf{UMGF} model cannot be improved as well. 
Therefore, the under-trained textual baselines of previous work make the effectiveness of the images unclear and we show that some of the MNER models perform even weaker than our \textbf{BERT-CRF} model. 

\begin{table}[t!]
\small
\centering
\setlength\tabcolsep{2pt}
\begin{tabular}{l||cc|cc|cc}
\hlineB{4}
\multicolumn{1}{c||}{} & \multicolumn{2}{c|}{Twitter-15} & \multicolumn{2}{c|}{Twitter-17}  & \multicolumn{2}{c}{SNAP} \\
 \hline
\multirow{3}{*}{Approach} & \multicolumn{2}{c|}{\bf Eval} & \multicolumn{2}{c|}{\bf Eval} & \multicolumn{2}{c}{\bf Eval}\\
 & \multicolumn{2}{c|}{\bf Modal} & \multicolumn{2}{c|}{\bf Modal} & \multicolumn{2}{c}{\bf Modal}\\
 & {\textbf{T}} & {\textbf{I+T}} & {\textbf{T}} & {\textbf{I+T}} & {\textbf{T}} & {\textbf{I+T}}\\
\hline\hline
{\textbf{ITA-Random}} & - & 74.67 & - & 84.98 & - & 85.82 \\
 {\textbf{ITA-GA$_{\text{BU}}$}} & - & 75.10 & - & 85.77 & - & 86.51 \\
{\textbf{ITA-LA$_{\text{BU}}$}} & - & 75.18 & - & 85.59  & - &  86.57\\
{\textbf{ITA-OCA$_{\text{Paddle}}$}} & - & 75.12 & - & \textbf{85.87}  & - & 86.66 \\
 {\textbf{BERT-CRF$_{\text{+ImgFeat}}$}} & - & 74.70 & - & 84.99 & - & 85.90\\
 {\textbf{VinVL-CRF}} & - & 60.58 & - & 75.55 & - & 74.53\\
 {\textbf{BERT+VinVL-CRF}} & - & 74.89 & - & 85.19 & - & 86.14 \\
 {\textbf{ITA-Joint}} & 74.88 & 75.22 & 85.31 & 85.60& 86.06 & 86.34  \\
\hline
 \multicolumn{7}{c}{\bf \textsc{References}}\\
\hline
 {\textbf{RpBERT w/o Rp}} & - & 72.60 & - & - & - & 86.20\\
 {\textbf{ITA-All$_{\text{+CVA}}$}} & \textbf{75.67} & \textbf{75.60} & \textbf{85.98} & 85.72 &  \textbf{86.83} & \textbf{86.75} \\
\hlineB{4}
\end{tabular}
\caption{A comparison of other variants of MNER models. }
\label{tab:comparison}
\end{table}

\subsection{Comparison with Other Variants}
To further show the effectiveness of ITA, we perform several comparisons between ITA and the following variants of the MNER model in Table \ref{tab:comparison}:
\paragraph{\textsc{\bf ITA-Random}:} We generate random image-text pairs for the model. For each sentence, we randomly select the image in the dataset and generate the corresponding visual contexts. The noises of random visual contexts make the model accuracy drop slightly comparing with our \textsc{\bf BERT-CRF} baseline, which shows the improvement of our approach is from the visual contexts rather than extending the input sequence length the embeddings.
\paragraph{\textsc{\bf ITA-Joint}:} It is an ablated model of \textsc{\bf ITA-All$_{\text{+CVA}}$}. We train the \textsc{\bf ITA-All} model for both input views without the CVA loss in Eq. \ref{eq:mda}. The model accuracy is improved moderately with only the \textbf{T} input view while our \textsc{\bf ITA-All$_{\text{+CVA}}$} can improve both input views significantly, which shows the effectiveness of the CVA module of ITA.
\paragraph{\textsc{\bf ITA-LA$_{\text{BU}}$} and \textsc{\bf ITA-GA$_{\text{BU}}$}:} We conduct experiments to see how the accuracy changes when using weaker image features. We use \textbf{B}ottom-\textbf{U}p features proposed by \citet{Anderson2017up-down} for object detection and image captioning. The captioning model is a pretrained image captioning model\footnote{\url{https://github.com/ruotianluo/self-critical.pytorch}} proposed by \citet{Luo2018DiscriminabilityOF} with the Bottom-Up features and self-critical training \citep{rennie2017self}. Results show that there is no significant difference between the visual contexts from Bottom-Up features and VinVL features. Therefore, our approaches can utilize other off-the-shelf vision models to extract visual contexts.
\paragraph{\textsc{\bf ITA-OCA$_{\text{Paddle}}$}:} We conduct experiments to see how the accuracy changes when using stronger OCR models. We use PaddleOCR\footnote{\url{https://github.com/PaddlePaddle/PaddleOCR}} for the experiment, which is one of the newest open resource lightweight OCR system. Results show that the model accuracy can be slightly improved comparing with \textsc{\bf ITA-OCA}, which shows the ITA models can be improved by using better OCR models.
\paragraph{{\textbf{BERT-CRF$_{\text{+ImgFeat}}$}}:} Instead of \textbf{ITA}, we can directly feed the image region features generated from an object detector into the BERT. We use ResNet-152 model to generate region features and then feed the features into a linear layer to project the region features into the same space of text features in the BERT. Moreover, we compare the model with {\textbf{RpBERT w/o Rp}}, which is an ablated model of {\textbf{RpBERT}} and is equivalent to {\textbf{BERT-CRF+$_{\text{+ImgFeat}}$}} over the usage of BERT embeddings. \citet{Sun2021RpBERTAT} showed {\textbf{RpBERT w/o Rp}} can improve the model accuracy compared with their baseline. However, our results show that the model accuracy slightly drops comparing with our {\textbf{BERT-CRF}}, which shows that it is difficult for the attention module of BERT to learn the relations of the unaligned representations of two modalities. 
\paragraph{\textsc{\bf VinVL-CRF}:} To show how the pretrained V+L models perform on the NER task, we use VinVL since it is a very recent state-of-the-art pretrained V+L model on a lot of multi-modal tasks. We feed the VinVL model with texts and images in the MNER datasets and finetune the model over the task. We take the text representations output from VinVL as the input of the CRF layer. The accuracy of the finetuned VinVL model drops significantly compared to the BERT model, which shows that the inductive bias of the pretrained V+L model hurts the model accuracy on MNER.
\paragraph{\textsc{\bf BERT+VinVL-CRF}:} As the VinVL model may lead to an inductive bias over the common nouns and the image, we jointly finetune the BERT and VinVL models and concatenate the output text representations of the two models. The accuracy is improved on a moderate scale, which shows BERT is complementary to VinVL for MNER.

\subsection{Analysis}

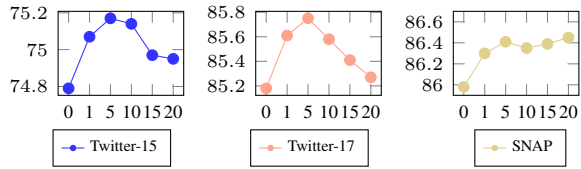
\begin{figure}[t!]
\begin{minipage}{1.0\linewidth}
\centering
\begin{tikzpicture}
    \begin{axis}[
        xshift=-10cm,
        name=ner,
        width=0.42\textwidth,
        height=0.35\textwidth,
        legend columns=2, 
        legend image post style={scale=0.5},
        legend style={font=\tiny,at={(0.9,-0.85)}, anchor=south east},
        tick label style={font=\scriptsize},
        xticklabels={0,1,5,10,15,20},
        xtick={0,1,2,3,4,5},
        ylabel style={font=\tiny,yshift=-0.2cm},
        ]
        \addplot[blue!80,mark=*] table[x=beam,y=acc] {twitter15.dat};
        \legend{Twitter-15}
    \end{axis}
    \begin{axis}[
        at={(ner.south west)},
        xshift=2.6cm,
        width=0.42\textwidth,
        height=0.35\textwidth,
        legend columns=2, 
        legend image post style={scale=0.5},
        legend style={font=\tiny,at={(0.9,-0.85)}, anchor=south east},
        tick label style={font=\scriptsize},
        xticklabels={0,1,5,10,15,20},
        xtick={0,1,2,3,4,5},
        ylabel style={font=\tiny,yshift=-0.5cm},
        ]
        \addplot[red!80!yellow!45,mark=*] table[x=beam,y=acc] {twitter17.dat};
        \legend{Twitter-17}
    \end{axis}
    \begin{axis}[
        at={(ner.south west)},
        xshift=5.2cm,
        width=0.42\textwidth,
        height=0.35\textwidth,
        legend columns=2, 
        legend image post style={scale=0.5},
        legend style={font=\tiny,at={(0.9,-0.85)}, anchor=south east},
        tick label style={font=\scriptsize},
        xticklabels={0,1,5,10,15,20},
        xtick={0,1,2,3,4,5},
        ylabel style={font=\tiny,yshift=-0.5cm},
        ymax=86.7,
        ymin=85.9,
        ]
        \addplot[yellow!80!blue!65,mark=*] table[x=beam,y=acc] {snap.dat};
        \legend{SNAP}
    \end{axis}
\end{tikzpicture}
\caption{A relation between the number of captions input to the MNER model and model accuracy. The x-axis is the number of captions. The y-axis is the averaged F1 score on the test set.}
\label{fig:test_curve}
\end{minipage}
\end{figure}

\paragraph{Effect of the Number of Captions}
Using more captions output from the captioning model can improve diversities of the visual contexts but can add noises to them as well. To better understand how the number of captions affects the model accuracy, we change the beam size and keep all the sentences output from the captioning model. The trends in Figure \ref{fig:test_curve} show that the model accuracy increases until $5$ captions for all the datasets and gradually drops when the number of captions further increases for Twitter-15 and 17 datasets. The observation shows that using $5$ captions keeps a good balance between the diversities and correctness of the captions.

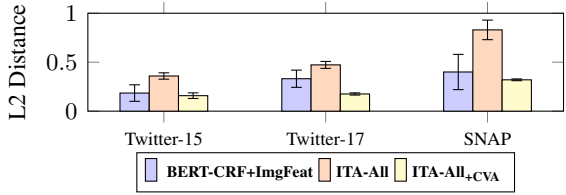
\begin{figure}[t]
\centering
\begin{tikzpicture}

\begin{axis}[
    ybar=0pt,
    bar width = {1em},
    width=0.49\textwidth,
    height=0.18\textwidth,
    enlarge x limits={abs=1cm},
    symbolic x coords={Twitter-15,Twitter-17, SNAP},
    xticklabel style={
         align=center, 
         font=\scriptsize, 
      },
    yticklabel style={font=\small},
    ylabel = {L2 Distance},
    ylabel style = {yshift=-0.3cm,font=\small},
    legend style={font=\tiny,at={(0.5,-0.83)},anchor=south},
    legend columns=3, 
    xtick={Twitter-15,Twitter-17, SNAP},
    legend image code/.code={%
      \draw[#1] (0cm,-0.1cm) rectangle (0.15cm,0.2cm);
    },
    ytick={0,0.5,1.0},
    ymin=0,
    ymax=1.0,
    ]
    \addplot[ybar, fill=blue!20, error bars/.cd,y dir=both,y explicit] table [x=type, y=base,y error=base_std] {mask_embed.dat};
    \addplot[ybar, fill=red!60!yellow!25, error bars/.cd,y dir=both,y explicit] table [x=type, y=ours,y error=ours_std] {mask_embed.dat};
    \addplot[ybar, fill=yellow!25, error bars/.cd,y dir=both,y explicit] table [x=type, y=ours-mda,y error=ours-mda_std] {mask_embed.dat};
    \legend{{\textbf{ BERT-CRF+ImgFeat}}, {\textbf{ITA-All}}, {\textbf{ITA-All$_{\text{+CVA}}$}}};
\end{axis}
\end{tikzpicture}
\caption{Averaged L2 distance between the token representations without image input ($\vr_i$) and with image input ($\vr^\prime_i$). The error bars mean the standard deviation over $5$ runs.}
\label{fig:l2dist}
\end{figure}

\paragraph{How ITA Eases the Cross-Modal Alignments}
Previous work such as \citet{moon-etal-2018-multimodal,Sun2021RpBERTAT} visualized modality attention in several cases to show the effectiveness of their approaches. However, visualizing the multi-layer attention in transformer-based embeddings is relatively difficult. Instead of studying special cases, we statistically calculate the averaged L2 distance between token representations $\vr_i$ and $\vr^{\prime}_i$ from two input modalities to show how the token representations depend on image information. In Figure \ref{fig:l2dist}, the L2 distance \textbf{ITA-All} is significantly larger than that of \textbf{BERT-CRF+ImgFeat}. Besides, the standard deviation of \textbf{BERT-CRF+ImgFeat} is very large. The observations show the image region features make the alignment become difficult and unstable while our visual contexts can significantly ease the cross-modal alignments. Moreover, with CVA, the L2 distance becomes much smaller and stable as CVA aligns the two input views to reduce the dependence on images, which shows the MNER model can better utilize the textual information with CVA.

\begin{table*}[t!]
\small
\centering
\begin{tabular}{l|ccc|ccc|ccc|ccc}
\hlineB{4}
 & \multicolumn{3}{c|}{LOC} & \multicolumn{3}{c|}{ORG} & \multicolumn{3}{c|}{PER} & \multicolumn{3}{c}{OTHER}\\
 & P & R & F1 & P & R & F1 & P & R & F1 & P & R & F1\\
\hline\hline
 & \multicolumn{12}{c}{Twitter-15}\\
 \hline
{\textbf{BERT-CRF}} & 80.0 & 83.8 & 81.8 & 65.9 & 61.0 & 63.3 & 84.2 & 86.8 & 85.4 & 44.2 & 44.2 & 44.1\\
{{\textbf{ITA-All$_{\text{+CVA}}$}}} & 81.1 & 84.2 & 82.6 & 68.8 & 60.6 & 64.4 & 84.0 & 87.2 & 85.6 & 44.9 & 44.6 & 44.8 \\
$\mathbf{\Delta}$ & 1.1  & 0.4  & 0.8  & 2.8  & -0.4 & 1.1  & -0.2 & 0.4  & 0.1  & 0.8  & 0.5  & 0.6 \\
\hline\hline
 & \multicolumn{12}{c}{Twitter-17} \\
\hline
{\textbf{BERT-CRF}} & 85.5 & 84.4 & 84.9 & 83.5 & 83.8 & 83.7 & 90.7 & 90.8 & 90.7 & 68.9 & 65.1 & 66.9 \\
{{\textbf{ITA-All$_{\text{+CVA}}$}}} & 86.0 & 83.7 & 84.8 & 83.9 & 84.2 & 84.0 & 91.9 & 90.9 & 91.4 & 73.7 & 64.3 & 68.6 \\
$\mathbf{\Delta}$ & 0.5  & -0.7 & -0.1 & 0.3  & 0.4  & 0.4  & 1.2  & 0.1  & 0.7  & 4.8  & -0.8 & 1.7 \\
\hline\hline
 & \multicolumn{12}{c}{SNAP} \\
\hline
{\textbf{BERT-CRF}} & 82.1 &82.8 &82.5 &87.8 &86.9 &87.3 &91.0 &91.5 &91.2 &72.3 &75.1 &73.7 \\
{{\textbf{ITA-All$_{\text{+CVA}}$}}} & 80.3 &81.7 &81.0 &87.8 &86.5 &87.1 &90.1 &91.2 &90.6 &70.1 &73.2 &71.6\\
$\mathbf{\Delta}$ & 1.9 &1.1 &1.5 &0.6 &0.5 &0.5 &0.9 &0.3 &0.6 &2.2 &1.9 &2.1 \\
\hlineB{4}
\end{tabular}
\caption{A comparison between our ITA ({\textbf{ITA-All$_{\text{+CVA}}$}} with \textbf{I+T} inputs) model and the baseline ({\textbf{BERT-CRF}}) in precision (P), recall (R) and F1. $\mathbf{\Delta}$ represents the relevant improvement of ITA over the Baseline.}
\label{tab:prf}
\end{table*}

\paragraph{How Images Affect the NER Prediction}
To study the effectiveness of the images over each label, we show a comparison between our model and our baselines in Table \ref{tab:prf}. When the relative improvement of the F1 score is larger than $0.5$, the relative improvement of precision is larger than that of recall. The observation shows that the main improvement of MNER is mainly because the images can help the model to reduce false-positive predictions for disambiguation on uncertain entities.\footnote{In Appendix \ref{app:case}, we show several cases to show the effectiveness of ITA to affect NER prediction.}

\section{Related Work}

\paragraph{Multi-modal Named Entity Recognition} 
Most of the previous approaches to MNER focus on the interaction between image and text features through attention mechanisms. \citet{moon-etal-2018-multimodal} proposed a modality attention network to fuse the text and image features before the input to the BiLSTM layer. \citet{lu-etal-2018-visual} additionally used a visual attention gate for the output features of the BiLSTM layer. \citet{zhang2018adaptive} proposed an adaptive co-attention network after the BiLSTM layer to model the interaction between image and text. Recently, \citet{10.1145/3394171.3413650} proposed OCSGA, which use object labels to model the interaction between image and object labels in an additional dense co-attention layer.
Compared with the work, we show a simpler and more effective way to utilize object labels and additionally use other alignment approaches to further improve the model accuracy. \citet{yu-etal-2020-improving-multimodal} proposed UMT, which utilized a multi-modal interaction module and an auxiliary entity span detection module for MNER. \citet{zhang2021multi} proposed UMGF, which utilizes a pretrained parser to create the graph connection between visual object tags and textual words. They used a graph attention network to fuse the textual and visual features. In order to better model whether the image is related to the text, \citet{Sun2021RpBERTAT} proposed RpBERT, which additionally trains on a text-image relation classification dataset proposed by \citet{vempala-preotiuc-pietro-2019-categorizing} to prevent the negative effect of noisy images. Comparing with RpBERT, we use CVA to let the NER model better utilize the input sentences without such kinds of supervision. All of these approaches focus on fusing the image and text features through the attention mechanism but ignore the gap between the image and text features while we propose to fully utilize the attention mechanism in the pretrained textual embeddings through aligning image features into textual space. Besides, some cross-media research also shows the effectiveness of OCR texts \citep{chen2016context,wang-etal-2020-cross-media} and object tags \citep{7780398} have been shown. Most of the approaches introduced a new attention module over cross-modal features while in comparison ITA effectively utilizes the attention module in the pretrained textual embeddings.

\paragraph{Pretrained Vision-Language Models}
Inspired by related work on language model pretraining, visual-language pretraining (VLP) has recently attracted a lot of attention \citep{li2019visualbert,Lu2019ViLBERTPT,chen2020uniter,Tan2019LXMERTLC,Li2020OscarOA,Yu2021ERNIEViLKE,zhang2021vinvl}. The pretrained V+L models are pretrained on large-scale image-text pairs and have achieved state-of-the-art accuracy over various vision-language tasks such as image captioning, VQA, NLVR and image-text retrieval. 
Recently, \citet{Li2020OscarOA} proposed Oscar to add object tags in pretraining so that self-attention can learn the image-text alignments easily. Following Oscar, \citet{zhang2021vinvl} proposed VinVL to train a large-scale object detector to improve the pretrained V+L model's accuracy. Comparing with VLP, MNER is a totally different task.
Firstly, the image-caption pairs are given in VLP and the image and text are equally important in pretraining for general representations. Therefore, using global alignment is meaningless for VLP but makes sense for MNER. In MNER, the input text is not the caption of the image and the image may not adds additional information to the input text.
Secondly, though captions and object tags are often utilized in VLP, how to effectively utilize the captions and object tags of the image in MNER is rarely considered. Finally, besides the local and global alignments, another aspect of ITA is the optical character alignment and cross-view alignment, which is rarely considered in VLP. 

\section{Conclusion}
In this paper, we propose Image-Text Alignments for multi-modal named entity recognition, which convert images into object labels, captions and OCR texts to align the image representations into textual space in a multi-level manner and form a cross-modal input view. The model can effectively utilize attention module of the transformer-based embeddings. Considering noises, availability of images and inference speed for practical use, we propose cross-view alignment, which let the MNER models better utilize the text information in the input. In our experiments, we show that ITA significantly outperforms previous state-of-the-art approaches on MNER datasets. We also show that most of the previous work failed to train a good textual baseline while our textual baseline can easily match or even outperform previous multi-modal approaches. In analysis, we further analyze how ITA eases the cross-modal alignments and how the images affect the NER prediction.

\section*{Acknowledgements}
This work was supported by the National Natural Science Foundation of China (61976139) and by Alibaba Group through Alibaba Innovative Research Program.

\bibliography{aaai22,anthology2,acl2021,custom}
\bibliographystyle{acl_natbib}
\appendix
\section{Appendix}

\begin{table*}[t!]
\small
\centering
\begin{tabular}{l||c|c|c}
\hlineB{4}
\multicolumn{1}{c||}{} & \multicolumn{1}{c|}{Twitter-15} & \multicolumn{1}{c|}{Twitter-17} & \multicolumn{1}{c}{SNAP} \\
\hline
Num Sents w/ OCR / Total Sents & 2,049 / 8,288 (24.72\%) & 1,197 / 4,461 (26.83\%) & 1,869 / 7,181 (26.03\%)\\
Avg. Length & 27.72 & 27.00 & 28.93\\
\hlineB{4}
\end{tabular}
\caption{A statistic about the number of sentences has OCR texts and the average length of OCR texts.}
\label{tab:ocr}
\end{table*}

\subsection{Details of Experiment Settings}
\label{app:setting}
We run our code on Tesla V100 GPU with 16 GB memory. It takes about two hours to train a model. The size of model parameter is approximately equal to size of BERT/XLMR embeddings. 

\subsection{Details of OCA}
\label{app:oca}
Table \ref{tab:ocr} shows that the OCR system only finds about 26\% sentences have texts in the image and the extracted texts have an average of 28 tokens. The statistics show that ITA-OCA can help to improve the model accuracy with only 26\% of the samples have OCR texts.

\begin{figure*}
	\centering
	\includegraphics[scale=0.5]{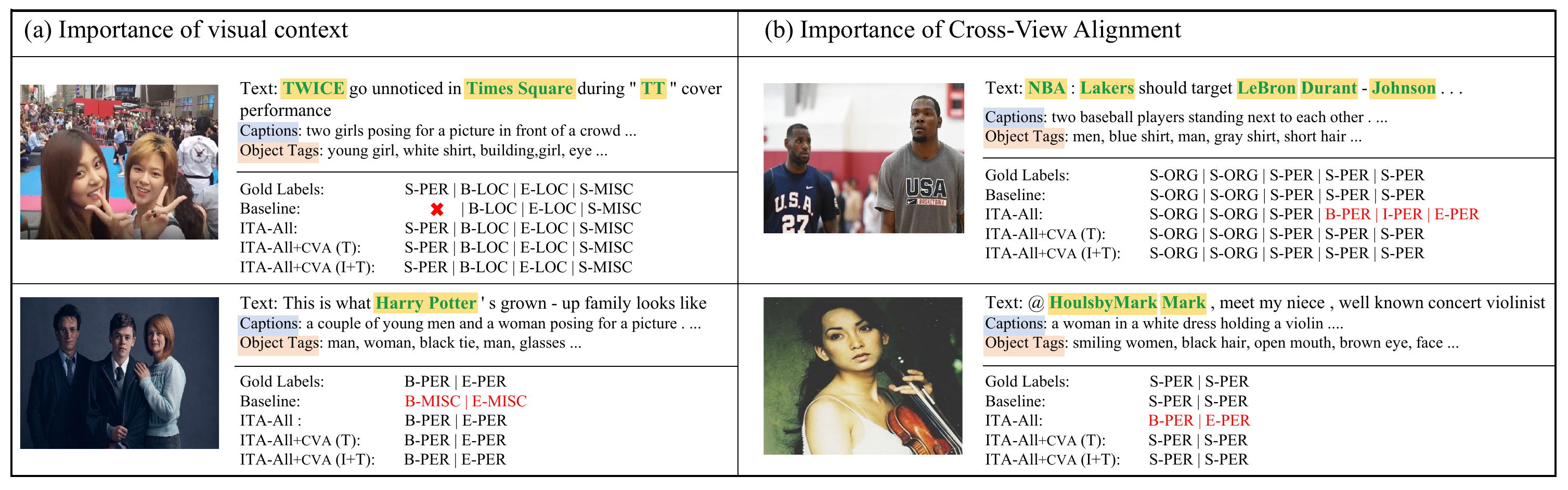}
	\caption{Examples of the positive and negative effects of images. The named entities in the text are colored. The wrongly predicted entities are marked in bold and colored in red. The missing entities are marked with \xmark. We use BIOES format to represent the label spans (\url{https://en.wikipedia.org/wiki/Inside-outside-beginning_(tagging)})}
	\label{fig:cases}
\end{figure*}

\begin{figure*}
	\centering
	\includegraphics[scale=1]{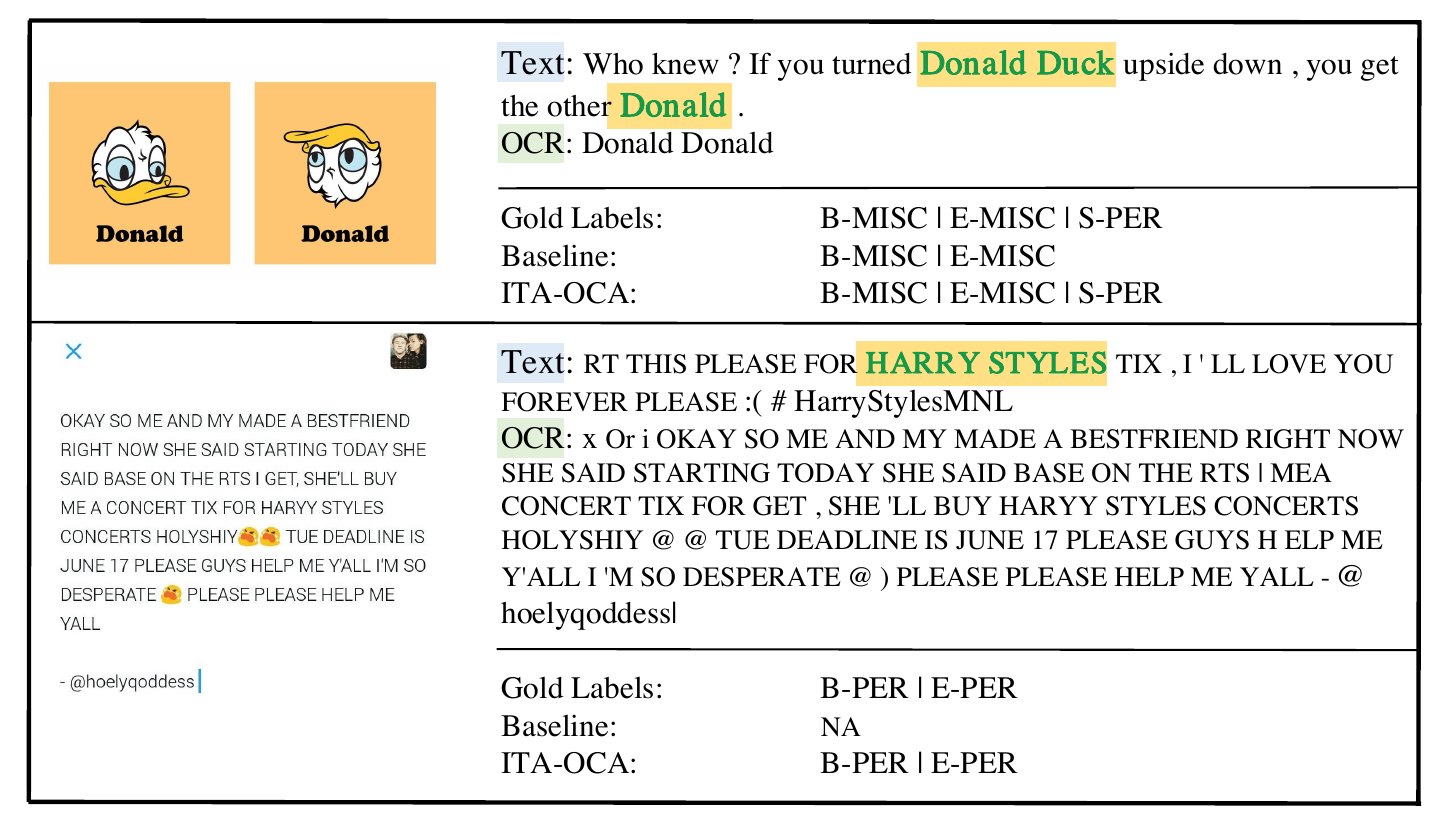}
	\caption{Examples of the positive effects of OCA. The named entities in the text are colored.}
	\label{fig:cases2}
\end{figure*}

\subsection{Case Study}
\label{app:case}
Despite that images can generally help to improve the accuracy of the NER model, there are a lot of cases that the images may contain misleading information to hurt the model prediction. We study two cases for LA nad GA: 1) the entities are wrongly predicted by \textbf{BERT-CRF} baseline but are correctly predicted by \textbf{ITA}; 2) the entities are wrongly predicted by \textbf{ITA} without CVA but are correctly predicted by the baseline and \textbf{ITA} with CVA. Figure \ref{fig:cases} shows the two cases with two samples for each. Figure \ref{fig:cases} (a) shows the first case, which shows the importance of the visual contexts. The baseline model failed to recognize the person entities ``TWICE'' and ``Harry Potter'' possibly because the two words are usually an adverb and a book name respectively. For the \textbf{I+T} input view, our MNER model is able to recognize the hints such as ``two girls'', ``young girl'', ``a couple of young men'' and ``woman'' in the visual contexts and then correctly predict the two entities. Figure \ref{fig:cases} (b) shows the second case, which shows how the noises from the image mislead the model predictions. There are three- and two-person entities in gold labels but the visual contexts indicate that the top right image has ``two baseball players'' and the bottom right image has only ``a woman''. As a result, \textbf{ITA} without CVA only predict two and one person entities according to the visual contexts in the two samples respectively. However, with CVA, \textbf{ITA} takes a good balance in utilizing the textual and visual information and correctly predicts the entity labels in both \textbf{T} and \textbf{I+T} input views.

For OCA, we study how the extracted texts can help model prediction. In the upper sample of Figure \ref{fig:cases2}, there are two ``Donald'' words in the image. The baseline model failed to identify the latter one while \textbf{ITA-OCA} can successfully identify both of them. In the bottom of Figure \ref{fig:cases2}, the texts in the image are mainly talking about ``HARRY STYLES'', which helps the model prediction.

\subsection{Discussion}
In our paper, we use the captioning and object detection model based on MSCOCO and visual genome. The model performance could be improved if we use domain-specific models (Twitter domain). For OCA, the model accuracy may be poor if the OCR system does not support a certain language.

\subsection{Loss Function Comparison with UMT}
\label{app:bug}
In the codes of UMT, the BERT embeddings tokenize the token in a sentence into subtokens. The codes use the first subtoken as the token representation to predict the corresponding label. However, for the other subtokens, the codes use a special label ``PAD'' for prediction. Therefore, the target labels are changed. For example, the original label sequence is ``B-X, I-X, O, B-X, O, O'' but now it becomes ``B-X, PAD, PAD, I-X, O, B-X, O, PAD, O''. As a result, the exact training objective changes compared with the training objective in the paper of UMT. We improve the code by removing all the ``PAD'' labels and just use the first subtoken of each token as the token representation. Our improved baseline model is significantly improved, while the accuracy of UMT model in the improved code cannot be further improved.
\end{document}

%% file: math_commands.tex
\usepackage{amsmath,amsfonts,bm}

\def\eqref#1{equation~\ref{#1}}

\def\1{\bm{1}}

\def\va{{\bm{a}}}

\def\vo{{\bm{o}}}

\def\vr{{\bm{r}}}

\def\vw{{\bm{w}}}
\def\vx{{\bm{x}}}
\def\vy{{\bm{y}}}

\def\evo{{o}}

\def\evy{{y}}

\DeclareMathAlphabet{\mathsfit}{\encodingdefault}{\sfdefault}{m}{sl}
\SetMathAlphabet{\mathsfit}{bold}{\encodingdefault}{\sfdefault}{bx}{n}













%% file: acl_latex.bbl
\begin{thebibliography}{47}
\expandafter\ifx\csname natexlab\endcsname\relax\def\natexlab#1{#1}\fi

\bibitem[{Agrawal et~al.(2015)Agrawal, Lu, Antol, Mitchell, Zitnick, Parikh,
  and Batra}]{Agrawal2015VQAVQ}
Aishwarya Agrawal, Jiasen Lu, Stanislaw Antol, Margaret Mitchell, C.~L.
  Zitnick, Devi Parikh, and Dhruv Batra. 2015.
\newblock Vqa: Visual question answering.
\newblock \emph{International Journal of Computer Vision}, 123:4--31.

\bibitem[{Akbik et~al.(2019)Akbik, Bergmann, and
  Vollgraf}]{akbik-etal-2019-pooled}
Alan Akbik, Tanja Bergmann, and Roland Vollgraf. 2019.
\newblock \href {https://doi.org/10.18653/v1/N19-1078} {Pooled contextualized
  embeddings for named entity recognition}.
\newblock In \emph{Proceedings of the 2019 Conference of the North {A}merican
  Chapter of the Association for Computational Linguistics: Human Language
  Technologies, Volume 1 (Long and Short Papers)}, pages 724--728, Minneapolis,
  Minnesota. Association for Computational Linguistics.

\bibitem[{Akbik et~al.(2018)Akbik, Blythe, and
  Vollgraf}]{akbik-etal-2018-contextual}
Alan Akbik, Duncan Blythe, and Roland Vollgraf. 2018.
\newblock \href {https://www.aclweb.org/anthology/C18-1139} {Contextual string
  embeddings for sequence labeling}.
\newblock In \emph{Proceedings of the 27th International Conference on
  Computational Linguistics}, pages 1638--1649, Santa Fe, New Mexico, USA.
  Association for Computational Linguistics.

\bibitem[{Anderson et~al.(2018)Anderson, He, Buehler, Teney, Johnson, Gould,
  and Zhang}]{Anderson2017up-down}
Peter Anderson, Xiaodong He, Chris Buehler, Damien Teney, Mark Johnson, Stephen
  Gould, and Lei Zhang. 2018.
\newblock Bottom-up and top-down attention for image captioning and visual
  question answering.
\newblock In \emph{CVPR}.

\bibitem[{Chen et~al.(2016)Chen, He, and Kan}]{chen2016context}
Tao Chen, Xiangnan He, and Min-Yen Kan. 2016.
\newblock Context-aware image tweet modelling and recommendation.
\newblock In \emph{Proceedings of the 24th ACM international conference on
  Multimedia}, pages 1018--1027.

\bibitem[{Chen et~al.(2020)Chen, Li, Yu, El~Kholy, Ahmed, Gan, Cheng, and
  Liu}]{chen2020uniter}
Yen-Chun Chen, Linjie Li, Licheng Yu, Ahmed El~Kholy, Faisal Ahmed, Zhe Gan,
  Yu~Cheng, and Jingjing Liu. 2020.
\newblock Uniter: Universal image-text representation learning.
\newblock In \emph{European conference on computer vision}.

\bibitem[{Conneau et~al.(2020)Conneau, Khandelwal, Goyal, Chaudhary, Wenzek,
  Guzm{\'a}n, Grave, Ott, Zettlemoyer, and
  Stoyanov}]{conneau-etal-2020-unsupervised}
Alexis Conneau, Kartikay Khandelwal, Naman Goyal, Vishrav Chaudhary, Guillaume
  Wenzek, Francisco Guzm{\'a}n, Edouard Grave, Myle Ott, Luke Zettlemoyer, and
  Veselin Stoyanov. 2020.
\newblock \href {https://doi.org/10.18653/v1/2020.acl-main.747} {Unsupervised
  cross-lingual representation learning at scale}.
\newblock In \emph{Proceedings of the 58th Annual Meeting of the Association
  for Computational Linguistics}, pages 8440--8451, Online. Association for
  Computational Linguistics.

\bibitem[{Derczynski et~al.(2017)Derczynski, Nichols, van Erp, and
  Limsopatham}]{derczynski-etal-2017-results}
Leon Derczynski, Eric Nichols, Marieke van Erp, and Nut Limsopatham. 2017.
\newblock \href {https://doi.org/10.18653/v1/W17-4418} {Results of the
  {WNUT}2017 shared task on novel and emerging entity recognition}.
\newblock In \emph{Proceedings of the 3rd Workshop on Noisy User-generated
  Text}, pages 140--147, Copenhagen, Denmark. Association for Computational
  Linguistics.

\bibitem[{Devlin et~al.(2019)Devlin, Chang, Lee, and
  Toutanova}]{devlin-etal-2019-bert}
Jacob Devlin, Ming-Wei Chang, Kenton Lee, and Kristina Toutanova. 2019.
\newblock \href {https://doi.org/10.18653/v1/N19-1423} {{BERT}: Pre-training of
  deep bidirectional transformers for language understanding}.
\newblock In \emph{Proceedings of the 2019 Conference of the North {A}merican
  Chapter of the Association for Computational Linguistics: Human Language
  Technologies, Volume 1 (Long and Short Papers)}, pages 4171--4186,
  Minneapolis, Minnesota. Association for Computational Linguistics.

\bibitem[{Do{\u{g}}an et~al.(2014)Do{\u{g}}an, Leaman, and Lu}]{dougan2014ncbi}
Rezarta~Islamaj Do{\u{g}}an, Robert Leaman, and Zhiyong Lu. 2014.
\newblock Ncbi disease corpus: a resource for disease name recognition and
  concept normalization.
\newblock \emph{Journal of biomedical informatics}, 47:1--10.

\bibitem[{Fetahu et~al.(2021)Fetahu, Fang, Rokhlenko, and
  Malmasi}]{10.1145/3404835.3463102}
Besnik Fetahu, Anjie Fang, Oleg Rokhlenko, and Shervin Malmasi. 2021.
\newblock \href {https://doi.org/10.1145/3404835.3463102} {Gazetteer enhanced
  named entity recognition for code-mixed web queries}.
\newblock In \emph{SIGIR '21}, SIGIR '21, New York, NY, USA. Association for
  Computing Machinery.

\bibitem[{He et~al.(2016)He, Zhang, Ren, and Sun}]{he2016deep}
Kaiming He, Xiangyu Zhang, Shaoqing Ren, and Jian Sun. 2016.
\newblock Deep residual learning for image recognition.
\newblock In \emph{Proceedings of the IEEE conference on computer vision and
  pattern recognition}, pages 770--778.

\bibitem[{Huang et~al.(2015)Huang, Xu, and Yu}]{Huang2015BidirectionalLM}
Zhiheng Huang, W.~Xu, and Kailiang Yu. 2015.
\newblock Bidirectional lstm-crf models for sequence tagging.
\newblock \emph{ArXiv}, abs/1508.01991.

\bibitem[{Lafferty et~al.(2001)Lafferty, McCallum, and
  Pereira}]{10.5555/645530.655813}
John~D. Lafferty, Andrew McCallum, and Fernando C.~N. Pereira. 2001.
\newblock Conditional random fields: Probabilistic models for segmenting and
  labeling sequence data.
\newblock In \emph{Proceedings of the Eighteenth International Conference on
  Machine Learning}, ICML ’01, page 282–289, San Francisco, CA, USA. Morgan
  Kaufmann Publishers Inc.

\bibitem[{Li et~al.(2016)Li, Sun, Johnson, Sciaky, Wei, Leaman, Davis,
  Mattingly, Wiegers, and Lu}]{li2016biocreative}
Jiao Li, Yueping Sun, Robin~J Johnson, Daniela Sciaky, Chih-Hsuan Wei, Robert
  Leaman, Allan~Peter Davis, Carolyn~J Mattingly, Thomas~C Wiegers, and Zhiyong
  Lu. 2016.
\newblock Biocreative v cdr task corpus: a resource for chemical disease
  relation extraction.
\newblock \emph{Database: The Journal of Biological Databases and Curation},
  2016.

\bibitem[{Li et~al.(2019)Li, Yatskar, Yin, Hsieh, and Chang}]{li2019visualbert}
Liunian~Harold Li, Mark Yatskar, Da~Yin, Cho-Jui Hsieh, and Kai-Wei Chang.
  2019.
\newblock Visualbert: A simple and performant baseline for vision and language.
\newblock \emph{arXiv preprint arXiv:1908.03557}.

\bibitem[{Li et~al.(2020{\natexlab{a}})Li, Yin, Li, Hu, Zhang, Zhang, Wang, Hu,
  Dong, Wei, Choi, and Gao}]{Li2020OscarOA}
Xiujun Li, Xi~Yin, Chunyuan Li, Xiaowei Hu, Pengchuan Zhang, Lei Zhang, Lijuan
  Wang, Houdong Hu, Li~Dong, Furu Wei, Yejin Choi, and Jianfeng Gao.
  2020{\natexlab{a}}.
\newblock Oscar: Object-semantics aligned pre-training for vision-language
  tasks.
\newblock In \emph{ECCV}.

\bibitem[{Li et~al.(2020{\natexlab{b}})Li, Yin, Li, Zhang, Hu, Zhang, Wang, Hu,
  Dong, Wei et~al.}]{li2020oscar}
Xiujun Li, Xi~Yin, Chunyuan Li, Pengchuan Zhang, Xiaowei Hu, Lei Zhang, Lijuan
  Wang, Houdong Hu, Li~Dong, Furu Wei, et~al. 2020{\natexlab{b}}.
\newblock Oscar: Object-semantics aligned pre-training for vision-language
  tasks.
\newblock In \emph{European Conference on Computer Vision}, pages 121--137.
  Springer.

\bibitem[{Lin et~al.(2014)Lin, Maire, Belongie, Hays, Perona, Ramanan,
  Doll{\'a}r, and Zitnick}]{lin2014microsoft}
Tsung-Yi Lin, Michael Maire, Serge Belongie, James Hays, Pietro Perona, Deva
  Ramanan, Piotr Doll{\'a}r, and C~Lawrence Zitnick. 2014.
\newblock Microsoft coco: Common objects in context.
\newblock In \emph{European conference on computer vision}, pages 740--755.
  Springer.

\bibitem[{Loshchilov and Hutter(2018)}]{loshchilov2018decoupled}
Ilya Loshchilov and Frank Hutter. 2018.
\newblock Decoupled weight decay regularization.
\newblock In \emph{International Conference on Learning Representations}.

\bibitem[{Lu et~al.(2018)Lu, Neves, Carvalho, Zhang, and
  Ji}]{lu-etal-2018-visual}
Di~Lu, Leonardo Neves, Vitor Carvalho, Ning Zhang, and Heng Ji. 2018.
\newblock \href {https://doi.org/10.18653/v1/P18-1185} {Visual attention model
  for name tagging in multimodal social media}.
\newblock In \emph{Proceedings of the 56th Annual Meeting of the Association
  for Computational Linguistics (Volume 1: Long Papers)}, pages 1990--1999,
  Melbourne, Australia. Association for Computational Linguistics.

\bibitem[{Lu et~al.(2019)Lu, Batra, Parikh, and Lee}]{Lu2019ViLBERTPT}
Jiasen Lu, Dhruv Batra, Devi Parikh, and Stefan Lee. 2019.
\newblock Vilbert: Pretraining task-agnostic visiolinguistic representations
  for vision-and-language tasks.
\newblock In \emph{NeurIPS}.

\bibitem[{Luo et~al.(2018)Luo, Price, Cohen, and
  Shakhnarovich}]{Luo2018DiscriminabilityOF}
R.~Luo, Brian~L. Price, Scott~D. Cohen, and Gregory Shakhnarovich. 2018.
\newblock Discriminability objective for training descriptive captions.
\newblock \emph{2018 IEEE/CVF Conference on Computer Vision and Pattern
  Recognition}, pages 6964--6974.

\bibitem[{Moon et~al.(2018)Moon, Neves, and
  Carvalho}]{moon-etal-2018-multimodal}
Seungwhan Moon, Leonardo Neves, and Vitor Carvalho. 2018.
\newblock \href {https://doi.org/10.18653/v1/N18-1078} {Multimodal named entity
  recognition for short social media posts}.
\newblock In \emph{Proceedings of the 2018 Conference of the North {A}merican
  Chapter of the Association for Computational Linguistics: Human Language
  Technologies, Volume 1 (Long Papers)}, pages 852--860, New Orleans,
  Louisiana. Association for Computational Linguistics.

\bibitem[{Rennie et~al.(2017)Rennie, Marcheret, Mroueh, Ross, and
  Goel}]{rennie2017self}
Steven~J Rennie, Etienne Marcheret, Youssef Mroueh, Jerret Ross, and Vaibhava
  Goel. 2017.
\newblock Self-critical sequence training for image captioning.
\newblock In \emph{Proceedings of the IEEE conference on computer vision and
  pattern recognition}, pages 7008--7024.

\bibitem[{Schweter and Akbik(2020)}]{schweter2020flert}
Stefan Schweter and Alan Akbik. 2020.
\newblock Flert: Document-level features for named entity recognition.
\newblock \emph{arXiv preprint arXiv:2011.06993}.

\bibitem[{Smith(2007)}]{smith2007overview}
Ray Smith. 2007.
\newblock An overview of the tesseract ocr engine.
\newblock In \emph{Ninth international conference on document analysis and
  recognition (ICDAR 2007)}, volume~2, pages 629--633. IEEE.

\bibitem[{Strauss et~al.(2016)Strauss, Toma, Ritter, de~Marneffe, and
  Xu}]{strauss-etal-2016-results}
Benjamin Strauss, Bethany Toma, Alan Ritter, Marie-Catherine de~Marneffe, and
  Wei Xu. 2016.
\newblock \href {https://www.aclweb.org/anthology/W16-3919} {Results of the
  {WNUT}16 named entity recognition shared task}.
\newblock In \emph{Proceedings of the 2nd Workshop on Noisy User-generated Text
  ({WNUT})}, pages 138--144, Osaka, Japan. The COLING 2016 Organizing
  Committee.

\bibitem[{Suhr et~al.(2019)Suhr, Zhou, Zhang, Bai, and Artzi}]{Suhr2019ACF}
Alane Suhr, Stephanie Zhou, Iris Zhang, Huajun Bai, and Yoav Artzi. 2019.
\newblock A corpus for reasoning about natural language grounded in
  photographs.
\newblock In \emph{ACL}.

\bibitem[{Sun et~al.(2020)Sun, Wang, Su, Weng, Sun, Zheng, and
  Chen}]{sun-etal-2020-riva}
Lin Sun, Jiquan Wang, Yindu Su, Fangsheng Weng, Yuxuan Sun, Zengwei Zheng, and
  Yuanyi Chen. 2020.
\newblock \href {https://www.aclweb.org/anthology/2020.coling-main.168}
  {{RIVA}: A pre-trained tweet multimodal model based on text-image relation
  for multimodal {NER}}.
\newblock In \emph{Proceedings of the 28th International Conference on
  Computational Linguistics}, pages 1852--1862, Barcelona, Spain (Online).
  International Committee on Computational Linguistics.

\bibitem[{Sun et~al.(2021)Sun, Wang, Zhang, Su, and Weng}]{Sun2021RpBERTAT}
Lin Sun, Jiquan Wang, Kai Zhang, Yindu Su, and Fangsheng Weng. 2021.
\newblock Rpbert: A text-image relation propagation-based bert model for
  multimodal ner.
\newblock In \emph{AAAI}.

\bibitem[{Sundheim(1995)}]{Sundheim1995NamedET}
Beth~M. Sundheim. 1995.
\newblock Named entity task definition, version 2.1.
\newblock In \emph{Proceedings of the Sixth Message Understanding Conference},
  pages 319--332.

\bibitem[{Tan and Bansal(2019)}]{Tan2019LXMERTLC}
Hao~Hao Tan and M.~Bansal. 2019.
\newblock Lxmert: Learning cross-modality encoder representations from
  transformers.
\newblock In \emph{EMNLP}.

\bibitem[{Tjong Kim~Sang(2002)}]{tjong-kim-sang-2002-introduction}
Erik~F. Tjong Kim~Sang. 2002.
\newblock \href {https://www.aclweb.org/anthology/W02-2024} {Introduction to
  the {C}o{NLL}-2002 shared task: Language-independent named entity
  recognition}.
\newblock In \emph{{COLING}-02: The 6th Conference on Natural Language Learning
  2002 ({C}o{NLL}-2002)}.

\bibitem[{Tjong Kim~Sang and
  De~Meulder(2003)}]{tjong-kim-sang-de-meulder-2003-introduction}
Erik~F. Tjong Kim~Sang and Fien De~Meulder. 2003.
\newblock \href {https://www.aclweb.org/anthology/W03-0419} {Introduction to
  the {C}o{NLL}-2003 shared task: Language-independent named entity
  recognition}.
\newblock In \emph{Proceedings of the Seventh Conference on Natural Language
  Learning at {HLT}-{NAACL} 2003}, pages 142--147.

\bibitem[{Vempala and
  Preo{\c{t}}iuc-Pietro(2019)}]{vempala-preotiuc-pietro-2019-categorizing}
Alakananda Vempala and Daniel Preo{\c{t}}iuc-Pietro. 2019.
\newblock \href {https://doi.org/10.18653/v1/P19-1272} {Categorizing and
  inferring the relationship between the text and image of {T}witter posts}.
\newblock In \emph{Proceedings of the 57th Annual Meeting of the Association
  for Computational Linguistics}, pages 2830--2840, Florence, Italy.
  Association for Computational Linguistics.

\bibitem[{Wang et~al.(2021)Wang, Jiang, Bach, Wang, Huang, Huang, and
  Tu}]{wang-etal-2021-improving}
Xinyu Wang, Yong Jiang, Nguyen Bach, Tao Wang, Zhongqiang Huang, Fei Huang, and
  Kewei Tu. 2021.
\newblock \href {https://doi.org/10.18653/v1/2021.acl-long.142} {Improving
  named entity recognition by external context retrieving and cooperative
  learning}.
\newblock In \emph{Proceedings of the 59th Annual Meeting of the Association
  for Computational Linguistics and the 11th International Joint Conference on
  Natural Language Processing (Volume 1: Long Papers)}, pages 1800--1812,
  Online. Association for Computational Linguistics.

\bibitem[{Wang et~al.(2020)Wang, Li, Lyu, and
  King}]{wang-etal-2020-cross-media}
Yue Wang, Jing Li, Michael Lyu, and Irwin King. 2020.
\newblock \href {https://doi.org/10.18653/v1/2020.emnlp-main.268} {Cross-media
  keyphrase prediction: A unified framework with multi-modality multi-head
  attention and image wordings}.
\newblock In \emph{Proceedings of the 2020 Conference on Empirical Methods in
  Natural Language Processing (EMNLP)}, pages 3311--3324, Online. Association
  for Computational Linguistics.

\bibitem[{Wu et~al.(2016)Wu, Shen, Liu, Dick, and Van Den~Hengel}]{7780398}
Qi~Wu, Chunhua Shen, Lingqiao Liu, Anthony Dick, and Anton Van Den~Hengel.
  2016.
\newblock \href {https://doi.org/10.1109/CVPR.2016.29} {What value do explicit
  high level concepts have in vision to language problems?}
\newblock In \emph{2016 IEEE Conference on Computer Vision and Pattern
  Recognition (CVPR)}, pages 203--212.

\bibitem[{Wu et~al.(2020)Wu, Zheng, Cai, Chen, Leung, and
  Li}]{10.1145/3394171.3413650}
Zhiwei Wu, Changmeng Zheng, Yi~Cai, Junying Chen, Ho-fung Leung, and Qing Li.
  2020.
\newblock Multimodal representation with embedded visual guiding objects for
  named entity recognition in social media posts.
\newblock ACM MM.

\bibitem[{Yamada et~al.(2020)Yamada, Asai, Shindo, Takeda, and
  Matsumoto}]{yamada-etal-2020-luke}
Ikuya Yamada, Akari Asai, Hiroyuki Shindo, Hideaki Takeda, and Yuji Matsumoto.
  2020.
\newblock \href {https://doi.org/10.18653/v1/2020.emnlp-main.523} {{LUKE}: Deep
  contextualized entity representations with entity-aware self-attention}.
\newblock In \emph{Proceedings of the 2020 Conference on Empirical Methods in
  Natural Language Processing (EMNLP)}, pages 6442--6454, Online. Association
  for Computational Linguistics.

\bibitem[{Young et~al.(2014)Young, Lai, Hodosh, and
  Hockenmaier}]{Young2014FromID}
Peter Young, Alice Lai, M.~Hodosh, and J.~Hockenmaier. 2014.
\newblock From image descriptions to visual denotations: New similarity metrics
  for semantic inference over event descriptions.
\newblock \emph{Transactions of the Association for Computational Linguistics},
  2:67--78.

\bibitem[{Yu et~al.(2021)Yu, Tang, Yin, Sun, Tian, Wu, and
  Wang}]{Yu2021ERNIEViLKE}
Fei Yu, Jiji Tang, Weichong Yin, Yu~Sun, Hao Tian, Hua Wu, and Haifeng Wang.
  2021.
\newblock Ernie-vil: Knowledge enhanced vision-language representations through
  scene graph.
\newblock In \emph{AAAI}.

\bibitem[{Yu et~al.(2020)Yu, Jiang, Yang, and
  Xia}]{yu-etal-2020-improving-multimodal}
Jianfei Yu, Jing Jiang, Li~Yang, and Rui Xia. 2020.
\newblock \href {https://doi.org/10.18653/v1/2020.acl-main.306} {Improving
  multimodal named entity recognition via entity span detection with unified
  multimodal transformer}.
\newblock In \emph{Proceedings of the 58th Annual Meeting of the Association
  for Computational Linguistics}, pages 3342--3352, Online. Association for
  Computational Linguistics.

\bibitem[{Zhang et~al.(2021{\natexlab{a}})Zhang, Wei, Li, Wu, Zhu, and
  Zhou}]{zhang2021multi}
Dong Zhang, Suzhong Wei, Shoushan Li, Hanqian Wu, Qiaoming Zhu, and Guodong
  Zhou. 2021{\natexlab{a}}.
\newblock Multi-modal graph fusion for named entity recognition with targeted
  visual guidance.
\newblock In \emph{Proceedings of the AAAI Conference on Artificial
  Intelligence}.

\bibitem[{Zhang et~al.(2021{\natexlab{b}})Zhang, Li, Hu, Yang, Zhang, Wang,
  Choi, and Gao}]{zhang2021vinvl}
Pengchuan Zhang, Xiujun Li, Xiaowei Hu, Jianwei Yang, Lei Zhang, Lijuan Wang,
  Yejin Choi, and Jianfeng Gao. 2021{\natexlab{b}}.
\newblock Vinvl: Revisiting visual representations in vision-language models.
\newblock In \emph{Proceedings of the IEEE/CVF Conference on Computer Vision
  and Pattern Recognition}, pages 5579--5588.

\bibitem[{Zhang et~al.(2018)Zhang, Fu, Liu, and Huang}]{zhang2018adaptive}
Qi~Zhang, Jinlan Fu, Xiaoyu Liu, and Xuanjing Huang. 2018.
\newblock Adaptive co-attention network for named entity recognition in tweets.
\newblock In \emph{Thirty-Second AAAI Conference on Artificial Intelligence}.

\end{thebibliography}
